\documentclass[letter]{article}

\usepackage[doublespacing]{setspace}
\usepackage{bm}
\usepackage{amsmath}
\usepackage{amssymb}
\usepackage{amsfonts}
\usepackage{graphicx}
\usepackage{algorithm}
\usepackage{algorithmic}
\def\vec#1{\mathbf{#1}}
\usepackage{comment}
\usepackage[margin=30mm]{geometry}

\begin{document}

\title{Meta-Learning for Koopman Spectral Analysis\\with Short Time-series}

\author{Tomoharu Iwata\\NTT Communication Science Laboratories, NTT Corporation \and Yoshinobu Kawahara\\Institute of Mathematics for Industry, Kyushu University\\
Center for Advanced Intelligence Project, RIKEN}

\date{}

\maketitle

\begin{abstract}
Koopman spectral analysis has attracted attention for nonlinear dynamical systems since we can analyze nonlinear dynamics with a linear regime by embedding data into a Koopman space by a nonlinear function. For the analysis, we need to find appropriate embedding functions. Although several neural network-based methods have been proposed for learning embedding functions, existing methods require long time-series for training neural networks. This limitation prohibits performing Koopman spectral analysis in applications where only short time-series are available. In this paper, we propose a meta-learning method for estimating embedding functions from unseen short time-series by exploiting knowledge learned from related but different time-series. With the proposed method, a representation of a given short time-series is obtained by a bidirectional LSTM for extracting its properties. The embedding function of the short time-series is modeled by a neural network that depends on the time-series representation. By sharing the LSTM and neural networks across multiple time-series, we can learn common knowledge from different time-series while modeling time-series-specific embedding functions with the time-series representation. Our model is trained such that the expected test prediction error is minimized with the episodic training framework. We experimentally demonstrate that the proposed method achieves better performance in terms of eigenvalue estimation and future prediction than existing methods.
\end{abstract}

\section{Introduction}

Analyzing nonlinear dynamical systems
is an important task in a wide variety of fields,
such as physics~\cite{braun1998nonlinear}, 
epidemiology~\cite{liu1987dynamical},
sociology~\cite{guastello2013chaos}, 
neuroscience~\cite{bullmore2009complex}, and
biology~\cite{daniels2015efficient}.
Recently, the Koopman operator theory~\cite{koopman1931hamiltonian,mezic2005spectral}  
has attracted attention for such analysis.
Based on the theory, a nonlinear dynamical system is lifted to the corresponding linear one
in a possibly infinite-dimensional space by embedding states using a nonlinear function.
Therefore, various methods developed for analyzing and controlling 
linear dynamical systems can be extended to nonlinear~\cite{morton2018deep,li2019learning}.
For example, frequencies and growth rates of nonlinear dynamical systems 
can be identified with the eigenvalues of a Koopman operator in the lifted space.

For data-driven Koopman spectral analysis,
we need appropriate functions to embed data in a finite-dimensional subspace, 
where the functions span a function space invariant to the Koopman operator.
We call this subspace the Koopman space.
One of the most popular algorithms for Koopman spectral analysis is dynamic mode decomposition (DMD)~\cite{schmid2010dynamic,rowley2009spectral,kutz2016dynamic}.
Since DMD assumes that data are obtained in a Koopman space,
appropriate embedding functions must be defined manually.
To automatically learn embedding functions from data,
several methods based on neural networks have been proposed~\cite{takeishi2017learning,lusch2018deep,yeung2019learning,lee2020model,azencot2020forecasting}.
Although these methods have shown to achieve high performance 
due to the high representation learning capability of neural networks,
they require long time-series for training.
The limitation prohibits performing 
Koopman spectral analysis 
in applications where only short time-series are available. 

In this paper, we propose a meta-learning method 
for estimating functions that embed data into a Koopman space.
By exploiting knowledge learned from related but different time-series,
the proposed method 
can estimate the embedding functions from short time-series.
In some applications, we can obtain multiple time-series,
e.g., fluid dynamics with different environments and 
electrocardiograph time-series from different people.
The proposed method enables us to perform 
Koopman spectral analysis
with short time-series obtained 
from new fluid environments or
unseen people.

Our model calculates a vector representation of
the given short time-series, which we call the time-series representation, 
by a bidirectional LSTM~\cite{hochreiter1997long}. 
The time-series representation contains 
the properties on the nonlinear dynamical system that generates 
the given short time-series. 
We model embedding functions by a feed-forward neural network 
that takes a data point and the time-series representation as input.
By using the time-series representation, 
the embedding functions are modified depending on the given short time-series,
and we can adapt the embedding functions 
to various nonlinear dynamical systems with
a single feed-forward neural network.

\begin{figure}[t!]
    \centering
    \includegraphics[width=23em]{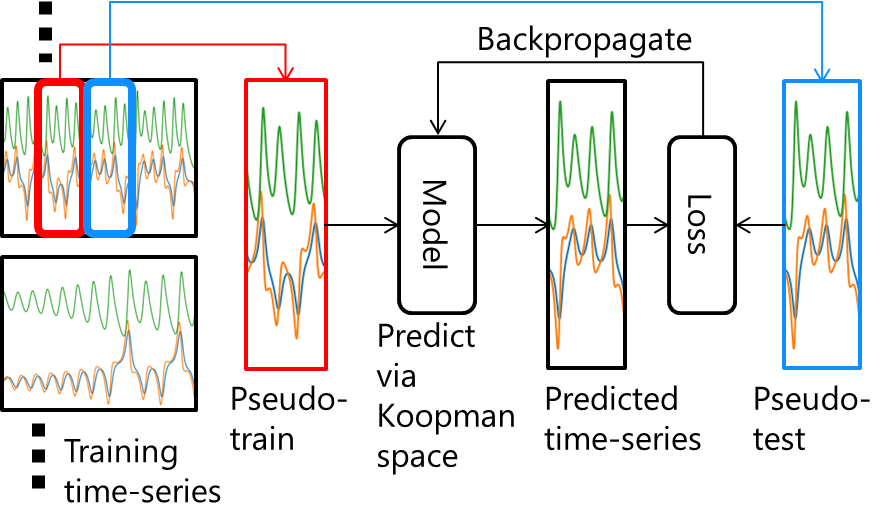}
    \caption{Our training framework. Multiple time-series are given for training. For each training epoch, pseudo-train and pseudo-test time-series are randomly sampled from training time-series. Our model predicts future time-series via a Koopman space from the pseudo-train time-series. As a loss, the error between the predicted and pseudo-test time-series is calculated. The parameters in our model are updated to minimize the prediction error by backpropagating the loss.}
    \label{fig:train}
\end{figure}

The neural networks in our model are trained 
such that the expected test prediction error is minimized 
with the episodic training framework~\cite{ravi2016optimization,santoro2016meta,snell2017prototypical,finn2017model,li2019episodic}, where the test phase is simulated
by randomly generating pseudo-training and pseudo-test data from multiple training time-series.
Figure~\ref{fig:train} illustrates the training procedure 
with the proposed method.
Since the neural networks in our model are shared across different time-series,
we can learn common knowledge from different time-series 
that can be useful in estimating embedding functions 
for various short time-series.
In addition, we can estimate embedding functions without retraining the neural networks when a new short time-series is given.

The following are the main contributions of this paper:
1) We propose the first meta-learning method for estimating nonlinear embedding functions for Koopman spectral analysis.
2) The proposed method can estimate embedding functions without retraining given new short time-series.
3) We experimentally confirm that the proposed method performs better than existing methods on eigenvalue estimation of Koopman operator and future prediction.
The remainder of this paper is organized as follows.
In Section~\ref{sec:related},
we briefly describe related work.
In Section~\ref{sec:preliminaries},
we explain Koopman operator theory,
on which our proposed method is based.
In Section~\ref{sec:proposed}, we propose a meta-learning method
for Koopman spectral analysis.
In Section~\ref{sec:experiments}, 
we experimentally 
demonstrate the effectiveness of the proposed method.
Finally, we present concluding remarks and discuss future work in Section~\ref{sec:conclusion}.

\section{Related work}
\label{sec:related}

With DMD, we need to manually prepare appropriate embedding functions 
according to the underlying nonlinear dynamics.
For modeling embedding functions, several methods have been proposed,
such as basis functions~\cite{williams2015data}
and kernels~\cite{williams2015kernel,kawahara2016dynamic}.
These methods work well only if appropriate basis functions or kernels are prepared.
For learning embedding functions from data,
several neural network-based methods have been
proposed~\cite{takeishi2017learning,yeung2019learning,lee2020model,lusch2018deep}.
However, they require long time-series for training.

There have been proposed many methods,
such as transfer learning, domain adaptation, multi-task learning~\cite{tan2018survey,long2017deep,killian2017robust,jia2018transfer,kumagai2019transfer}, and meta-learning~\cite{schmidhuber:1987:srl,bengio1991learning,ravi2016optimization,andrychowicz2016learning,vinyals2016matching,snell2017prototypical,bartunov2018few,finn2017model,qin2019recurrent,kim2019attentive,willi2019recurrent,garnelo2018conditional},
to improve performance using data from related but different domains
for regression, classification, and time-series prediction~\cite{hooshmand2019energy,ribeiro2018transfer,lemke2010meta,prudencio2004meta,talagala2018meta,ali2018cross,oreshkin2019n,oreshkin2020meta,iwata2020fewshot}.
However, these methods are inapplicable for Koopman spectral analysis.
The proposed method obtains a task (time-series) representation 
using neural networks, which is similar to encoder-decoder 
style meta-learning methods~\cite{garnelo2018neural,kumagai2018zero,iwata2020fewshot,xu2019metafun}, 
such as neural processes~\cite{garnelo2018conditional}. The encoder-decoder style methods obtain a task representation by an encoder, 
and output a task-specific prediction by a decoder using the task representation.
On the other hand, the proposed method uses
the task (time-series) representation for obtaining
task-specific embedding functions for Koopman spectral analysis.

\section{Preliminaries: Koopman operator theory}
\label{sec:preliminaries}

We consider a nonlinear discrete-time dynamical system:
\begin{align}
    \vec{x}_{t+1}=f(\vec{x}_{t}),
\end{align}
where $\vec{x}_{t}\in\mathcal{X}$ is a state at timestep $t$.
Koopman operator $\mathcal{K}$ is defined as an infinite-dimensional linear operator that acts on observables $g:\mathcal{X}\rightarrow\mathbb{R}$ (or $\mathbb{C}$)~\cite{koopman1931hamiltonian}:
\begin{align}
    g(\vec{x}_{t+1})=\mathcal{K}g(\vec{x}_{t}),
\end{align}
with which the analysis of nonlinear dynamics
can be lifted to a linear (but infinite-dimensional) regime.
When $\mathcal{K}$ has only discrete spectra,
observable $g$ is expanded by the eigenfunctions of $\mathcal{K}$,
$\varphi_{d}:\mathcal{X}\rightarrow\mathbb{C}$:
\begin{align}
g(\vec{x}_{t})=\sum_{d=1}^{\infty}\alpha_{d}\varphi_{d}(\vec{x}_{t}),
\end{align}
where $\alpha_{d}\in\mathbb{C}$ is the coefficient.
Then the dynamics of observable $g$ is expanded:
\begin{align}
g(\vec{x}_{t})=(g\circ \underbrace{f \circ \cdots \circ f}_{t-1})(\vec{x}_{1})=\sum_{d=1}^{\infty}\lambda_{d}^{t-1}\alpha_{d}\varphi_{d}(\vec{x}_{1}),
\label{eq:koopman}
\end{align}
where $\circ$ is the composition, and $\lambda_{d}\in\mathbb{C}$ is the eigenvalue of eigenfunction $\varphi_{d}$.
Since $\lambda_{d}$ is the only time dependent factor in the right-hand side of Eq.~(\ref{eq:koopman}),
$\lambda_{d}$ characterizes the time evolution, i.e.,
its phase determines the frequency and its magnitude determines the growth rate.

Although the existence of the Koopman operator is theoretically guaranteed in various situations, its practical use is limited by its infinite dimensionality. 
We can assume the restriction of $\mathcal{K}$ 
to a finite-dimensional subspace $\mathcal{G}$~\cite{takeishi2017learning}. 
If $\mathcal{G}$ is spanned by a finite number of functions $\{g_{1},\dots,g_{D}\}$,
then the restriction of $\mathcal{K}$ to $\mathcal{G}$, 
which we denote $\vec{K}\in\mathbb{R}^{D\times D}$ and call the Koopman matrix, 
becomes a finite-dimensional operator, and
\begin{align}
    \vec{g}_{t+1}=\vec{K}\vec{g}_{t},
    \label{eq:gKg}
\end{align}
where $\vec{g}_{t}=[g_{1}(\vec{x}_{t}),\dots,g_{D}(\vec{x}_{t})]\in\mathbb{R}^{D}$ is a vector of observables at timestep $t$,
and which we call the Koopman embedding.

\section{Proposed method}
\label{sec:proposed}

\subsection{Problem formulation}

In the training phase,
we are given a set of time-series,
$\mathcal{D}=\{\vec{Y}_{1},\cdots,\vec{Y}_{D}\}$,
where 
$\vec{Y}_{d}=[\vec{y}_{d1},\cdots,\vec{y}_{dT_{d}}]$
is the time-series data of the $d$th task,
and $\vec{y}_{dt}\in\mathbb{R}^{M}$
is the measurement vector at timestep $t$.
In the test phase,
we are given a target short time-series 
$\vec{Y}_{*}=[\vec{y}_{*1},\cdots,\vec{y}_{*T_{*}}]$,
which we call the support time-series,
where $\vec{y}_{*t}\in\mathbb{R}^{M}$.
The support time-series is different from time-series
in training data $\mathcal{D}$.
Our aim is to estimate the Koopman matrix 
of the target time-series.

\subsection{Model}

\begin{figure}[t!]
    \centering
    \includegraphics[width=20em]{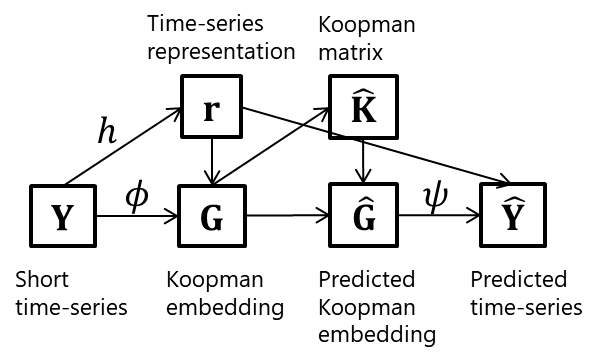}
    \caption{Our model. Short time-series $\vec{Y}_{\mathrm{S}}$ is given as input. First, time-series representation $\vec{r}_{\mathrm{S}}$ is calculated from $\vec{Y}_{\mathrm{S}}$ by bidirectional LSTM $h$. Second, Koopman embedding $\vec{G}_{\mathrm{S}}$ is obtained using $\vec{Y}_{\mathrm{S}}$ and $\vec{r}_{\mathrm{S}}$ by neural network $\phi$. Third, Koopman matrix $\vec{K}_{\mathrm{S}}$ is estimated by least squares using $\vec{G}_{\mathrm{S}}$. Forth, future Koopman embeddings $\hat{\vec{G}}_{\mathrm{S}}$ are predicted using $\vec{K}_{\mathrm{S}}$. Fifth, future time-series $\hat{\vec{Y}}_{\mathrm{S}}$ are predicted by mapping $\hat{\vec{G}}_{\mathrm{S}}$ into the measurement space by neural network $\psi$.} 
    \label{fig:model}
\end{figure}

Our model predicts future measurement vectors 
via an estimated Koopman matrix
given support time-series $\vec{Y}_{\mathrm{S}}=[\vec{y}_{\mathrm{S}1},\cdots,\vec{y}_{\mathrm{S}T}]$.
Figure~\ref{fig:model} illustrates our model.
In the training phase, support time series 
$\vec{Y}_{\mathrm{S}}$ is randomly generated from given set of time-series $\mathcal{D}$ as described in Section~\ref{sec:training}.
In the test phase, it corresponds $\vec{Y}_{*}$ described in the previous subsection.
Subscript $\mathrm{S}$ represents that the variable is task dependent, where given support set $\vec{Y}_{\mathrm{S}}$ is obtained from the task.

First, we obtain a time-series representation that represents the properties of support time-series $\vec{Y}_{\mathrm{S}}$
using bidirectional LSTM.
Let $\vec{r}^{\mathrm{F}}_{\mathrm{S}t}\in\mathbb{R}^{K}$
and $\vec{r}^{\mathrm{B}}_{\mathrm{S}t}\in\mathbb{R}^{K}$
be forward and backward hidden states
of the support time-series at timestep $t$
calculated by
\begin{align}
  \vec{r}^{\mathrm{F}}_{\mathrm{S}t}=h^{\mathrm{F}}(\vec{r}^{\mathrm{F}}_{\mathrm{S},t-1},\vec{y}_{\mathrm{S}t}),\quad
  \vec{r}^{\mathrm{B}}_{\mathrm{S}t}=h^{\mathrm{B}}(\vec{r}^{\mathrm{B}}_{\mathrm{S},t+1},\vec{y}_{\mathrm{S}t}),
\end{align}
where $h^{\mathrm{F}}$ and $h^{\mathrm{B}}$ are forward and backward LSTMs.
The forward (backword) hidden state $\vec{r}^{\mathrm{F}}_{\mathrm{S}t}$ ($\vec{r}^{\mathrm{B}}_{\mathrm{S}t}$)
contains
information about the time-series before (after) timestep $t$.
We calculate time-series representation $\vec{r}_{\mathrm{S}}\in\mathbb{R}^{2K}$
by averaging concatenated forward and backward hidden states:
\begin{align}
    \vec{r}_{\mathrm{S}}=\frac{1}{T}\sum_{t=1}^{T}[\vec{r}^{\mathrm{F}}_{\mathrm{S}t},\vec{r}^{\mathrm{B}}_{\mathrm{S}t}],
\end{align}
where $[\cdot,\cdot]$ represents vector concatenation.

Second, each measurement vector is embedded in the Koopman space using time-series representation $\vec{r}$:
\begin{align}
    \vec{g}_{\mathrm{S}t}=\phi([\vec{y}_{\mathrm{S}t},\vec{r}_{\mathrm{S}}]),
    \label{eq:g}
\end{align}
where $\phi$ is a feed-forward neural network.
By inputting the time-series representation,
we can adapt Koopman embedding $\vec{g}_{\mathrm{S}t}$ to support time-series $\vec{Y}_{\mathrm{S}}$.
In some applications, 
we cannot necessarily observe full states $\vec{x}_{\mathrm{S}t}$.
By measurement function $\phi:\mathbb{R}^{M+2K}\rightarrow\mathbb{R}^{D}$,
we can obtain Koopman embedding $\vec{g}_{\mathrm{S}t}$ from measurement vector $\vec{y}_{\mathrm{S}t}$
without explicitly estimating state $\vec{x}_{\mathrm{S}t}$.

Third, we estimate Koopman matrix $\vec{K}_{\mathrm{S}}$ using Koopman embeddings 
$\vec{g}_{\mathrm{S}1},\dots,\vec{g}_{\mathrm{S}T}$.
We define two matrices, $\vec{G}_{\mathrm{S}1}$ and 
$\vec{G}_{\mathrm{S}2}$, which are constructed from the Koopman embeddings:
\begin{align}
  \vec{G}_{\mathrm{S}1}=[\vec{g}_{\mathrm{S}1},\vec{g}_{\mathrm{S}2},
    \cdots,\vec{g}_{\mathrm{S},T-1}]\in\mathbb{R}^{D\times (T-1)},
  \quad
  \vec{G}_{\mathrm{S}2}=[\vec{g}_{\mathrm{S}2},\vec{g}_{\mathrm{S}3},
    \cdots,\vec{g}_{\mathrm{S}T}]\in\mathbb{R}^{D\times (T-1)},  
  \label{eq:G}
\end{align}
where $\vec{G}_{\mathrm{S}1}$ consists of the Koopman embeddings from timestep 1 to $T-1$, 
and $\vec{G}_{\mathrm{S}2}$ consists of those at their next timesteps.
Using the two matrices,
the Koopman matrix is estimated in a closed form 
by minimizing the squared error
of the predicted Koopman embeddings in Eq.~(\ref{eq:gKg}):
\begin{align}
    \hat{\vec{K}}_{\mathrm{S}}&=\arg\min_{\vec{K}_{\mathrm{S}}}\parallel \vec{G}_{\mathrm{S}2}-\vec{K}_{\mathrm{S}}\vec{G}_{\mathrm{S}1}\parallel^{2}
            =\vec{G}_{\mathrm{S}2}\vec{G}_{\mathrm{S}1}^{\dagger},
\label{eq:Khat}            
\end{align}
where $\dagger$ is pseudo-inverse.
Since the pseudo-inverse is differentiable, we can backpropagate the loss
    through $\hat{\vec{K}}_{\mathrm{S}}$ to train neural networks.

Fourth, we predict the Koopman embedding at timestep $\tau$
from that at $T$, which is the last one in the support time-series,
by multiplying the estimated Koopman matrix $\tau-T$ times:
\begin{align}
\hat{\vec{g}}_{\mathrm{S}\tau}=\hat{\vec{K}}_{\mathrm{S}}^{\tau-T}\vec{g}_{\mathrm{S}T}.
\label{eq:ghat}
\end{align}

Fifth, the predicted Koopman embeddings are mapped into the measurement space by
\begin{align}
\hat{\vec{y}}_{\mathrm{S}\tau}=\psi([\hat{\vec{g}}_{\mathrm{S}\tau},\vec{r}_{\mathrm{S}}]),
\label{eq:yhat}
\end{align}
where $\psi:\mathbb{R}^{D+2K}\rightarrow\mathbb{R}^{M}$
is a feed-forward neural network,
and $\hat{\vec{y}}_{\tau}$ is a predicted measurement vector at timestep $\tau$.
We can adapt the mapping function 
using the time-series representation.
Our model can be seen as a single neural network
that takes support time-series $\vec{Y}_{\mathrm{S}}$ and timestep $\tau$ as input,
and outputs predicted measurement vector $\hat{\vec{y}}_{\mathrm{S}\tau}$ at timestep $\tau$.

\subsection{Training}
\label{sec:training}

We train parameters of neural networks in our model, $h^{\mathrm{F}}$, $h^{\mathrm{B}}$, $\phi$, $\psi$, by minimizing the expected test prediction error:
\begin{align}
    \hat{\bm{\Theta}}=\arg\min_{\bm{\Theta}}
    \mathbb{E}_{d\sim\{1,\dots,D\}}\Bigg[\mathbb{E}_{(\vec{Y}_{\mathrm{Q}},\vec{Y}_{\mathrm{S}})\sim\vec{Y}_{d}}\Bigg[
    L(\vec{Y}_{\mathrm{Q}},\vec{Y}_{\mathrm{S}}) 
    \Bigg]\Bigg],
    \label{eq:E}
\end{align}
where $\bm{\Theta}$ is a set of parameters of the neural networks,
$\mathbb{E}$ represents the expectation,
$\vec{Y}_{\mathrm{S}}$ is the support time-series 
that is sampled from training time-series $\vec{Y}_{d}$, 
$\vec{Y}_{\mathrm{Q}}$ is the query time-series with length $T_{\mathrm{Q}}$ that is the succeeding time-series of sampled support time-series $\vec{Y}_{\mathrm{S}}$ in $\vec{Y}_{d}$, 
\begin{align}
    L(\vec{Y}_{\mathrm{Q}},\vec{Y}_{\mathrm{S}}) = \frac{1}{T_{\mathrm{Q}}}\sum_{\tau=1}^{T_{\mathrm{Q}}}\parallel\hat{\vec{y}}_{\mathrm{Q},\tau}(\vec{Y}_{\mathrm{S}})-\vec{y}_{\mathrm{Q},\tau}\parallel^{2},
    \label{eq:L}
\end{align}
is the prediction error of the query time-series,
and
$\hat{\vec{y}}_{\mathrm{Q},\tau}(\vec{Y}_{\mathrm{S}})$ is the predicted measurement vector at timestep $\tau$ in query time-series $\vec{Y}_{\mathrm{Q}}$ by our model using support time-series $\vec{Y}_{\mathrm{S}}$.
The support time-series is a pseudo-train time-series, and the query time-series is a pseudo-test time-series in Figure~\ref{fig:train}.
Existing neural network-based methods for Koopman spectral analyis 
train neural networks by minimizing the prediction error
of the training data~\cite{takeishi2017learning,yeung2019learning,lee2020model,lusch2018deep}.
With the test error instead of the training error 
for our objective function,
we can improve the generalization performance.
The training procedure for our model
is shown in Algorithm~\ref{alg}.

\begin{algorithm}[t]
  \caption{Training procedure for our model.}
  \label{alg}
  \begin{algorithmic}[1]
    \renewcommand{\algorithmicrequire}{\textbf{Input:}}
    \renewcommand{\algorithmicensure}{\textbf{Output:}}
    \REQUIRE{Multipe training time-series $\mathcal{D}$,
      support time-series length $T$, 
      query time-series length $T_{\mathrm{Q}}$}
    \ENSURE{Trained neural network parameters $\bm{\Phi}$}
    \WHILE{not done}
    \STATE Randomly sample time-series $d$ from $\{1,\dots,D\}$.
    \STATE Randomly sample timestep $t$ from \\ $\{1,T_{d}-T-T_{\mathrm{Q}}\}$.
    \STATE Construct support time-series\\ $\vec{Y}_{\mathrm{S}}=[\vec{y}_{d,t},\dots,\vec{y}_{d,t+T-1}]$.
    \STATE Construct query time-series\\ $\vec{Y}_{\mathrm{Q}}=[\vec{y}_{d,t+T},\dots,\vec{y}_{d,t+T+T_{\mathrm{Q}}-1}]$.
    \STATE Predict query time-series from support time-series $\vec{Y}_{\mathrm{S}}$ by our model.
    \STATE Calculate loss by Eq.~(\ref{eq:L}), and its gradients.
    \STATE Update model parameters $\bm{\Phi}$ using the loss and gradients.
    \ENDWHILE
  \end{algorithmic}
\end{algorithm}

\section{Experiments}
\label{sec:experiments}

\subsection{Data}

We evaluated the proposed method using four datasets:
Synthetic, Van-der-Pol, Lorenz, and Cylinder-wake.

\begin{figure*}[t]
\centering
{\tabcolsep=0.1em \begin{tabular}{cccc}
\includegraphics[width=11em]{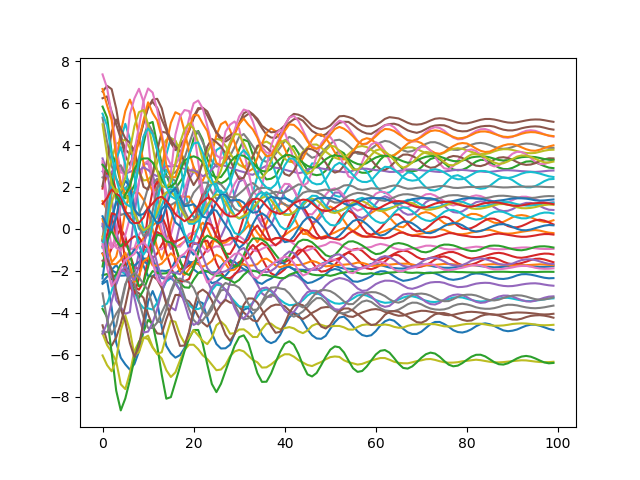}&
\includegraphics[width=11em]{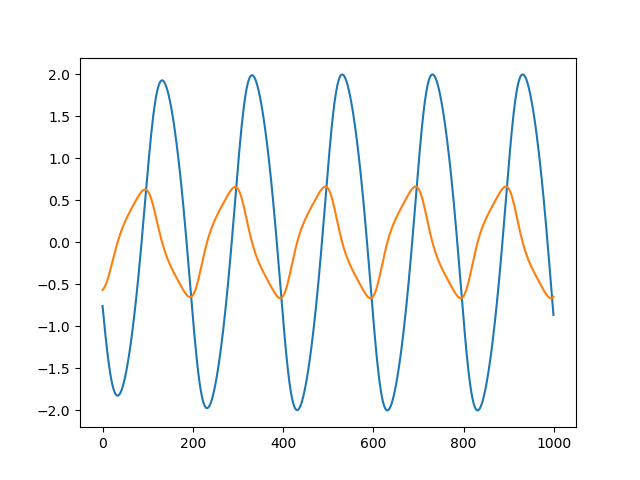}&
\includegraphics[width=11em]{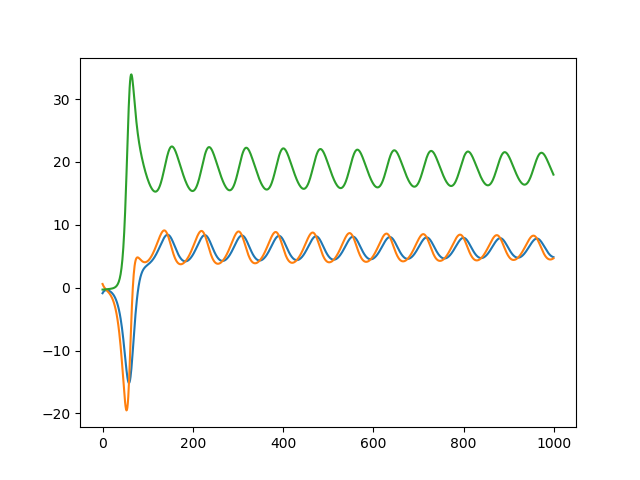}&
\includegraphics[width=11em]{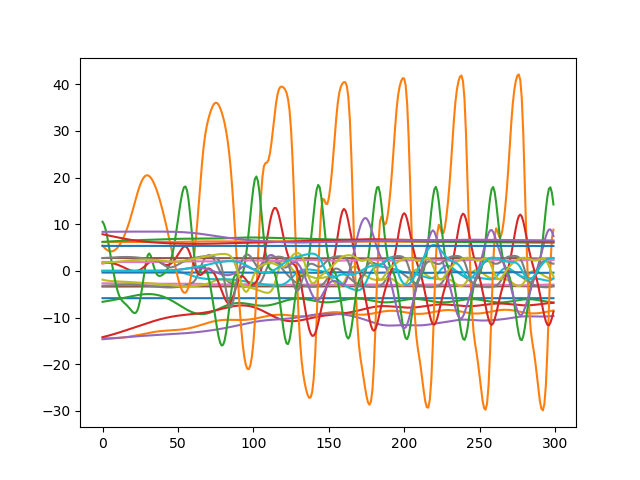}\\
\includegraphics[width=11em]{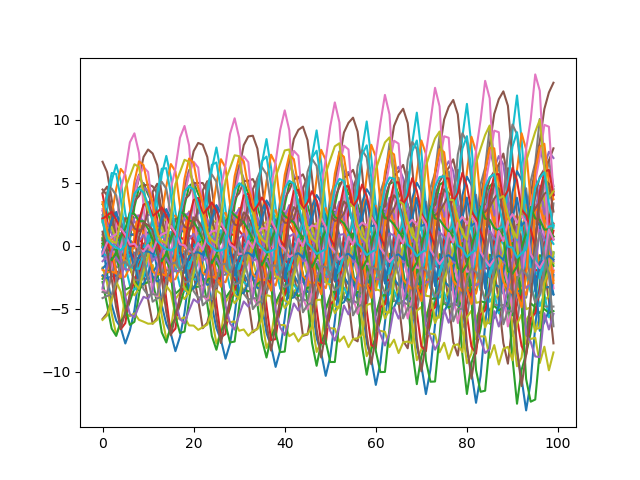}&
\includegraphics[width=11em]{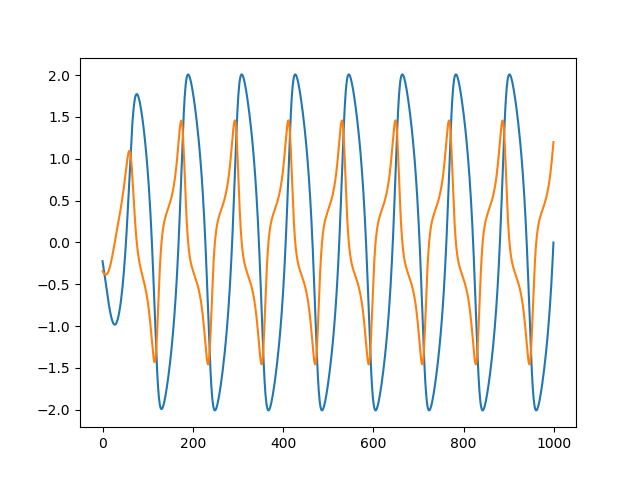}&
\includegraphics[width=11em]{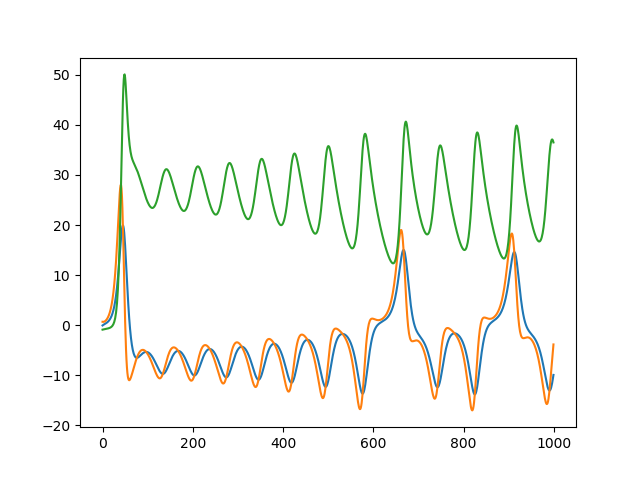}&
\includegraphics[width=11em]{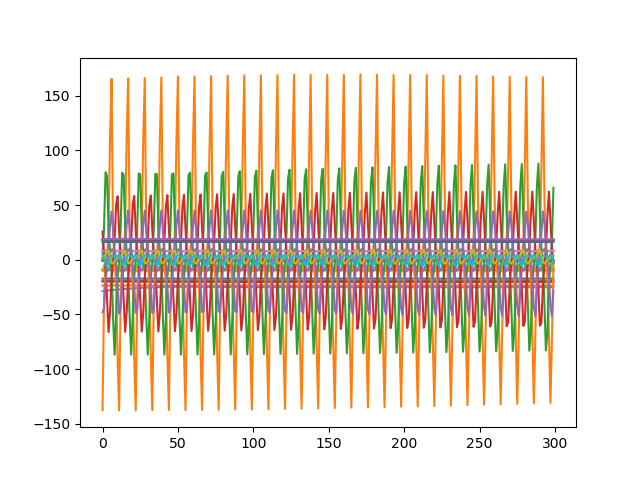}\\
\includegraphics[width=11em]{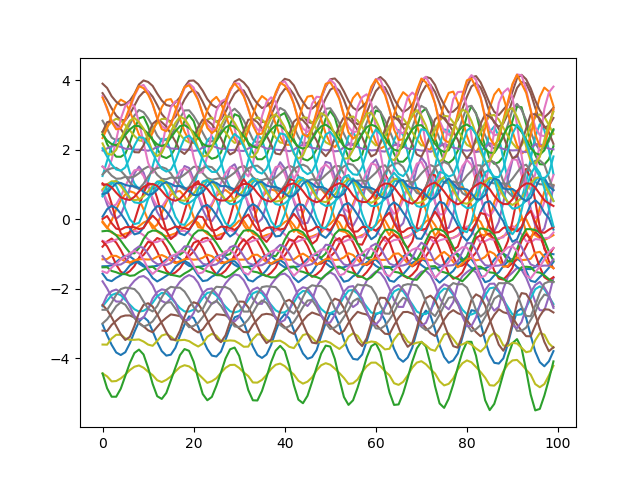}&
\includegraphics[width=11em]{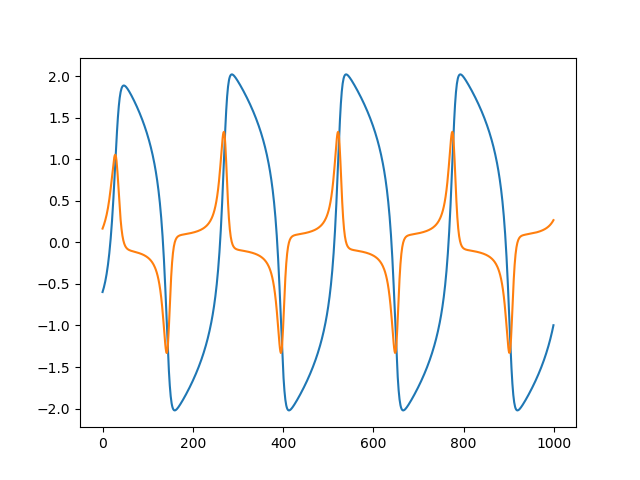}&
\includegraphics[width=11em]{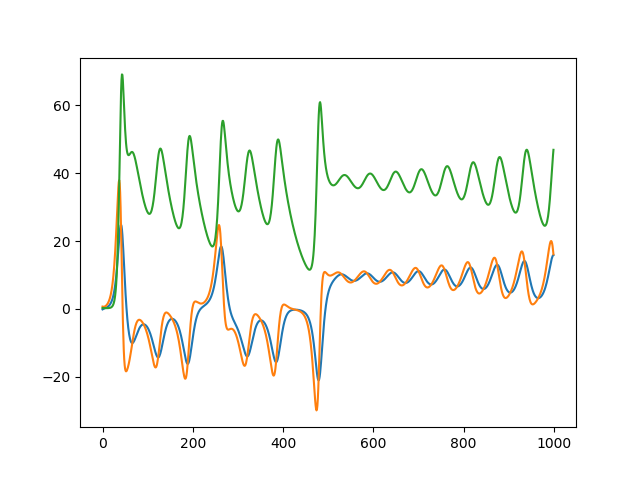}&
\includegraphics[width=11em]{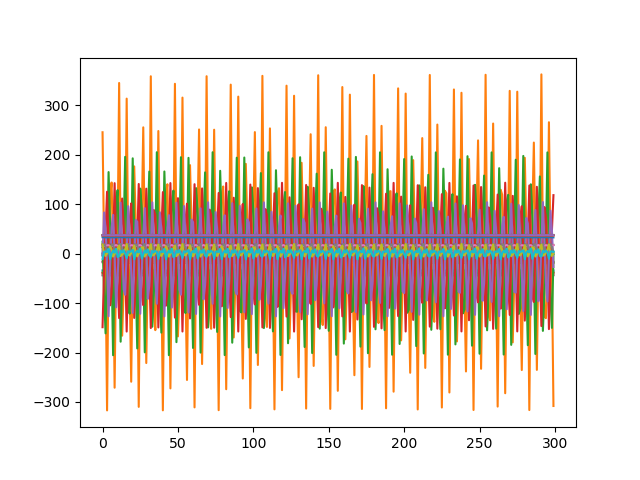}\\
(a) Synthetic & (b) Van-der-Pol & (c) Lorenz & (d) Cylinder-wake \\
\end{tabular}}
\caption{Examples of time-series in Synthetic, Van-der-Pol, Lorenz, and Cylinder-wake data used in our experiments.}
\label{fig:data}
\end{figure*}

Synthetic data were generated by the following procedure.
First, the dynamics of Koopman embeddings 
in a two-dimensional Koopman space was obtained
with a randomly generated linear transition matrix.
Then the dynamics in a ten-dimensional measurement space was obtained by a nonlinear transformation using a randomly generated 
three-layered feed-forward neural network. 
The neural network took the concatenation of a Koopman embedding 
and a randomly generated two-dimensional time-series representation.
We generated 81 time-series with different dynamics using different linear transition matrices and different time-series representations.
Figure~\ref{fig:data}(a) shows examples 
of the generated Synthetic time-series.

The Van-der-Pol data were generated using 
the van der Pol oscillator~\cite{van1926lxxxviii}, which is
the following non-conservative oscillator with nonlinear damping:
\begin{align}
    \frac{dy_{1}}{dt}=y_{2}, \quad
    \frac{dy_{2}}{dt}=ay_{2}(1-y_{1}^{2})-by_{1},
\end{align}
with a two-dimensional measurement space.
The van der Pol oscillator is used in many scientific fields,
e.g., as a model for action potentials of neurons~\cite{nagumo1962active,fitzhugh1969mathematical},
and a model for two plates in a geological fault in seismology~\cite{guckenheimer2003forced}.
We generated 100 time-series with different dynamics
using ten different $a\in[0.1,2]$ 
and ten different $b\in[0.1,2]$.
Figure~\ref{fig:data}(b) shows examples 
of the generated Van-der-Pol time-series.

The Lorenz data were generated using the Lorenz system~\cite{lorenz1963deterministic},
which has chaotic solutions and is developed for atmospheric convection:
\begin{align}
    \frac{dy_{1}}{dt}=\sigma(y_{2}-y_{1}), \quad
    \frac{dy_{2}}{dt}=y_{1}(\rho-y_{3})-y_{2}, \quad
    \frac{dy_{3}}{dt}=y_{1}y_{2}-\beta y_{3},
\end{align}
with a three-dimensional measurement space.
We generated 900 time-series with different dynamics
using 30 different $\rho\in[20,80]$,
30 different $\beta\in[2,5]$,
and fixed $\sigma=10$.
Figure~\ref{fig:data}(c) shows examples 
of the generated Lorenz time-series.

\begin{figure}[t]
\centering
\includegraphics[width=11em]{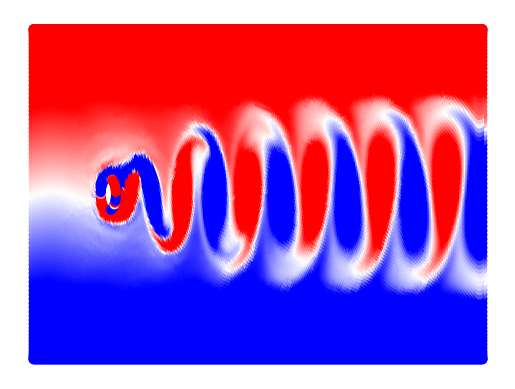}
\caption{Original Cylinder-wake data.}
\label{fig:cylinder}
\end{figure}

The Cylinder-wake data were generated 
using a fluid dynamics simulator whose exemplary snapshot is shown in Figure~\ref{fig:cylinder}.
We sampled 100 time-series with different dynamics
with 25 different sets of 25 grid locations
and four different Reynolds numbers $\{100,300,600,1000\}$.
Figure~\ref{fig:data}(d) shows examples 
of the generated Cylinder-wake time-series.

For all data, we used 70\% of the time-series for training, 10\% for validation, and the remainder for testing.
We performed 50 experiments with 
different train, validation and test splits, 
and evaluated their average.

\subsection{Proposed method setting}

We used bidirectional LSTM $h^{\mathrm{F}}$, $h^{\mathrm{B}}$
with $K=32$ hidden units.
For neural networks for the mapping 
between the measurement and Koopman space
$\phi$, $\psi$,
we used four-layered feed-forward neural networks with 128 hidden units.
The dimension of the Koopman space was $D=2$.
The activation function in the neural networks was rectified linear unit, $\mathrm{ReLU}(x)=\max(0,x)$.
Optimization was performed using Adam~\cite{kingma2014adam} 
with learning rate $10^{-3}$ and
dropout rate $0.1$.
The maximum number of training epochs was 10,000,
and the validation data were used for early stopping.
We set support time-series length at $T=20$,
and query time-series length at $T_{Q}=20$.

\subsection{Comparing methods}

We compared the proposed method with 
dynamic mode decomposition (DMD)~\cite{schmid2010dynamic,rowley2009spectral,kutz2016dynamic},
neural network-based DMD (NDMD)~\cite{takeishi2017learning},
finetuning of NDMD (Finetune), and
model-agnostic meta-learning~\cite{finn2017model} of NDMD (MAML).
For neural network-based methods, NDMD, Finetune, and MAML,
we used the neural networks with the same architecture with the proposed method.

With DMD, Koopman embeddings were defined by measurement vectors,
$\vec{g}_{t}=\vec{y}_{t}$,
and the Koopman matrix was estimated by Eq.~(\ref{eq:Khat})
using the measurement vector in the support time-series.

With NDMD, the neural network-based embedding functions 
between the measurement and Koopman spaces
were trained by minimizing the prediction error of the training time-series.
In particular,
Koopman embeddings were obtained by
\begin{align}
    \vec{g}_{t}=\phi_{\mathrm{N}}(\vec{y}_{t}),
    \label{eq:gN}
\end{align}
instead of Eq.~(\ref{eq:g}),
and 
predicted measurement vectors were obtained by
\begin{align}
    \hat{\vec{y}}_{t}=\psi_{\mathrm{N}}(\hat{\vec{g}}_{t}),
    \label{eq:yhatN}    
\end{align}
instead of Eq.~(\ref{eq:yhat}),
where $\phi_{\mathrm{N}}$ and 
$\psi_{\mathrm{N}}$ were feed-forward neural networks.
The neural networks were trained by minimizing the training prediction error:
\begin{align}
    \frac{1}{T}\sum_{\tau=1}^{T}\parallel\hat{\vec{y}}_{\tau}(\vec{Y})-\vec{y}_{\tau}\parallel^{2},
    \label{eq:Ls}
\end{align}
where $\hat{\vec{y}}_{\tau}(\vec{Y})$ was the predicted measurement vector at timestep $\tau$ 
in $\vec{Y}$ 
that was predicted using support-time series $\vec{Y}$.
NDMD did not use time-series representations.

With Finetune,
the neural networks in NDMD were finetuned
for each target time-series by minimizing the prediction error of the support time-series
in the test phase.
The number of epochs for finetuning was 1,000.

With MAML,
the initial parameters of neural networks in NDMD 
were trained using model-agnostic meta-learning with the episodic training framework,
such that the test predictive performance improves when finetuned.
The number of epochs for finetuning was five.

For the ablation study, we also compared with two versions of the proposed method: OursT and OursN.
OursT trained the neural networks by minimizing the training prediction error in Eq.~(\ref{eq:Ls}) instead of Eq.~(\ref{eq:L}). 
OursN was the proposed method without time-series representations,
where Eqs.~(\ref{eq:gN},\ref{eq:yhatN}) 
were used for mapping between the measurement and Koopman spaces,
and the neural networks were trained by minimizing
the test prediction error.

\subsection{Results}

\begin{table*}[!t]
    \centering
    {\tabcolsep=0.3em\begin{tabular}{rrrrrrr}
    \hline      
    Ours & DMD & NDMD & Finetune & MAML & OursT & OursN \\
    \hline      
    {\bf 0.399$\pm$0.012} & 44.855$\pm$8.391 & 0.740$\pm$0.008 & 0.777$\pm$0.002 & 0.689$\pm$0.013 & 0.408$\pm$0.015 & 0.809$\pm$0.008\\
        \hline
    \end{tabular}}
    \caption{Averaged eigenvalue estimation error and its standard error with the Synthetic data.}
    \label{tab:eigen}
\end{table*}

  \begin{table*}[t!]
  \centering
    \begin{tabular}{lrrrr}
    \hline      
      & Synthetic & Van-der-Pol & Lorenz & Cylinder-wake \\
    \hline      
Ours & {\bf 1.197$\pm$0.224} &{\bf 0.227$\pm$0.011} &{\bf 3.226$\pm$0.086} &{\bf 14.734$\pm$1.484} \\
DMD &  4532.802$\pm$106.014 & 4.280$\pm$0.406 &  605.178$\pm$24.009 & 1613.074$\pm$56.930 \\
NDMD & 143.895$\pm$64.756 & 0.551$\pm$0.019 & 70.096$\pm$16.909 & 62.243$\pm$18.986 \\
Finetune & 3.938$\pm$0.045 & 2.197$\pm$0.024 & 19.886$\pm$0.120 & 59.893$\pm$1.596 \\
MAML & 3.153$\pm$0.099 & 0.609$\pm$0.010 & 11.742$\pm$0.655 & 40.570$\pm$1.810 \\
OursT & 90.758$\pm$62.354 & 0.726$\pm$0.031 & 12.514$\pm$0.586 & 15.775$\pm$2.862 \\
OursN & 178.235$\pm$63.512& 0.398$\pm$0.009 & 55.654$\pm$18.310 & 41.471$\pm$1.643\\
    \hline
    \end{tabular}
    \caption{Averaged test prediction error and its standard error.}
\label{tab:rmse}
  \end{table*}

\begin{figure*}[t!]
\centering
{\tabcolsep=-0.2em \begin{tabular}{ccccc}
\includegraphics[width=9.5em]{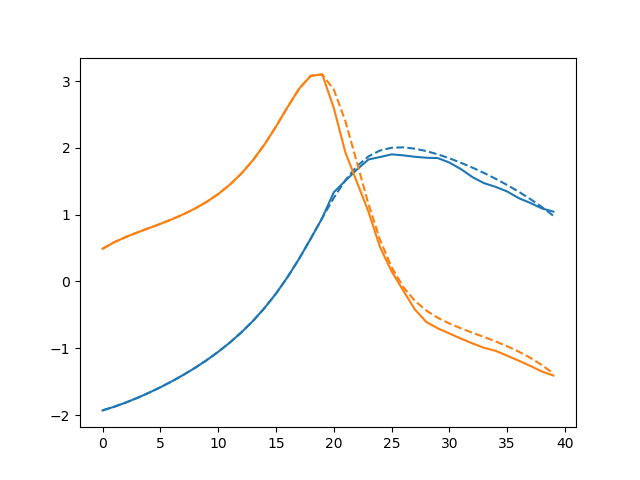}&
\includegraphics[width=9.5em]{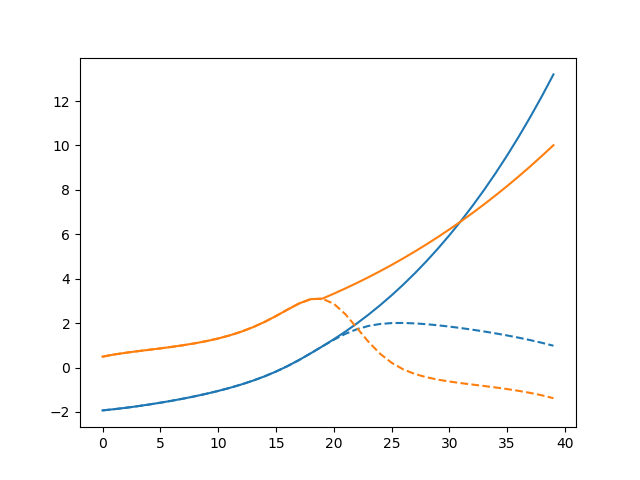}&
\includegraphics[width=9.5em]{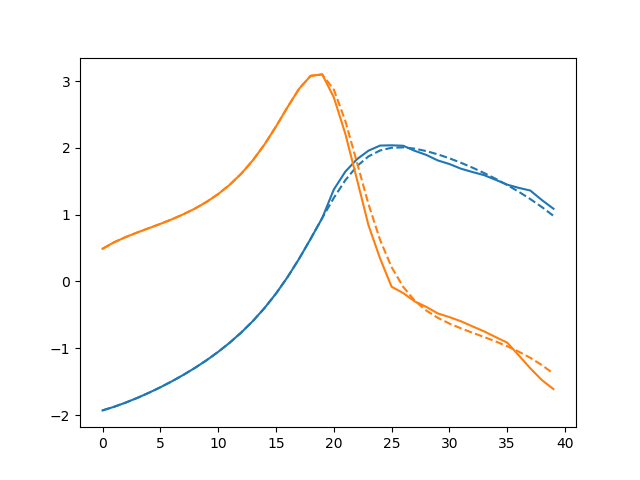}&
\includegraphics[width=9.5em]{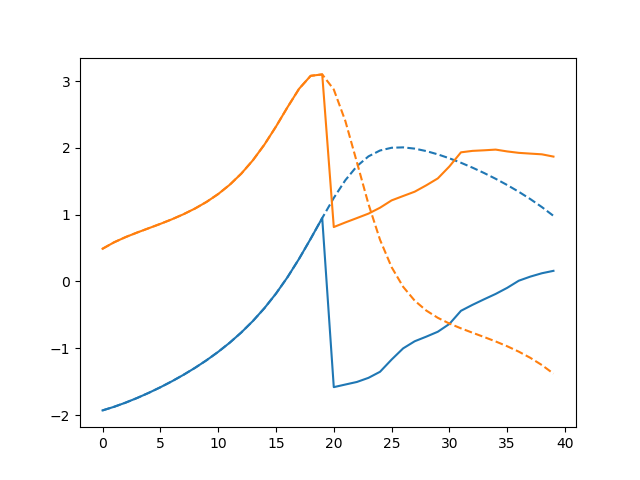}&
\includegraphics[width=9.5em]{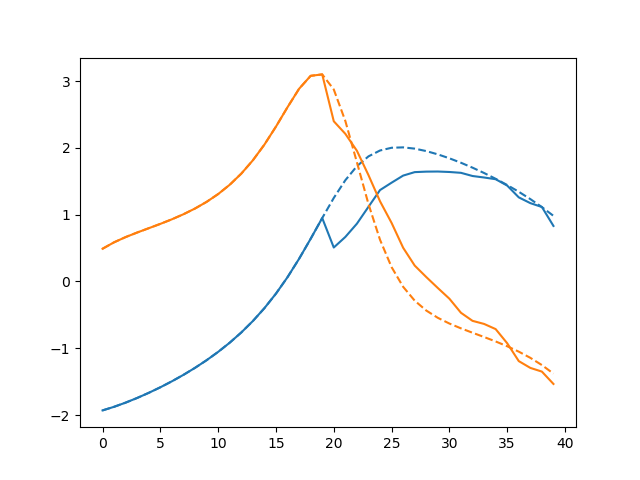}\\
\includegraphics[width=9.5em]{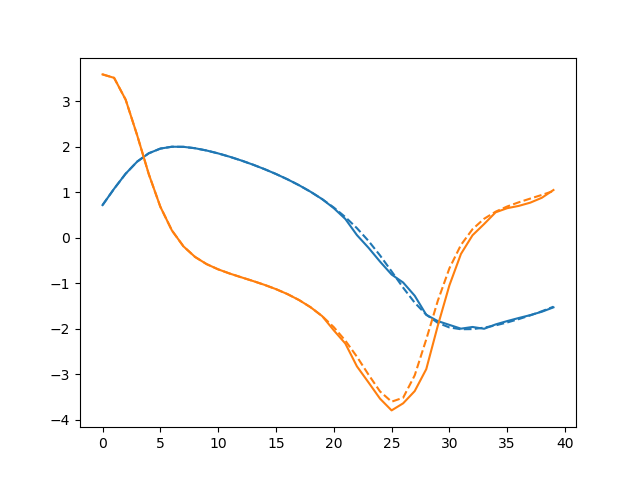}&
\includegraphics[width=9.5em]{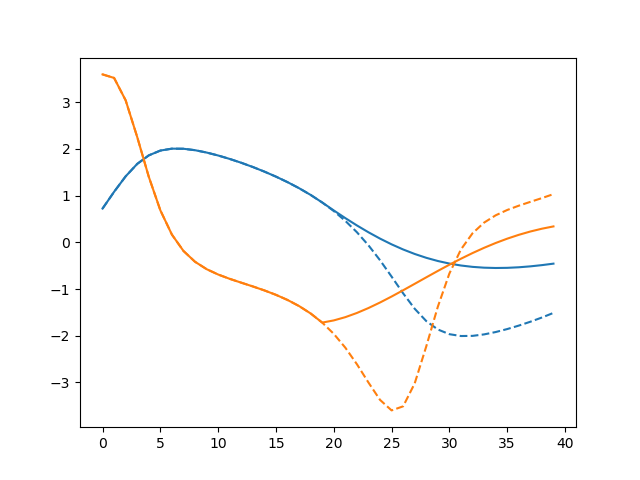}&
\includegraphics[width=9.5em]{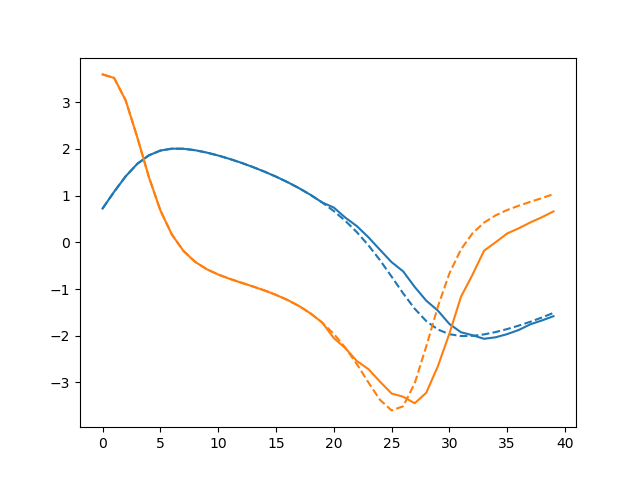}&
\includegraphics[width=9.5em]{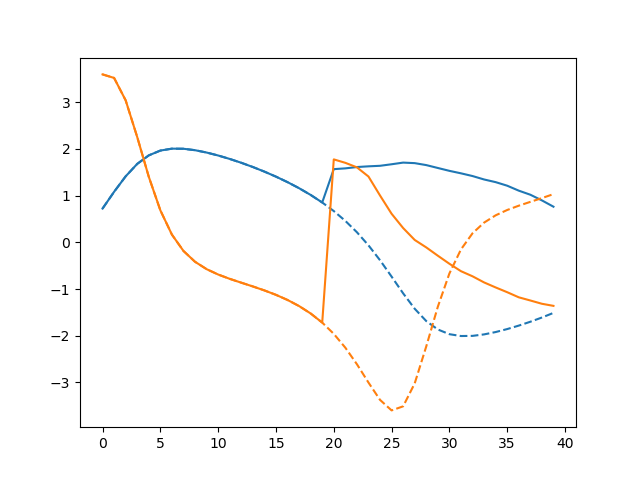}&
\includegraphics[width=9.5em]{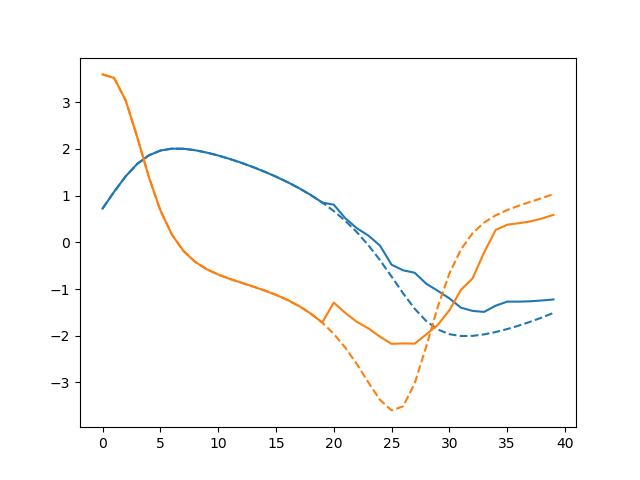}\\
\includegraphics[width=9.5em]{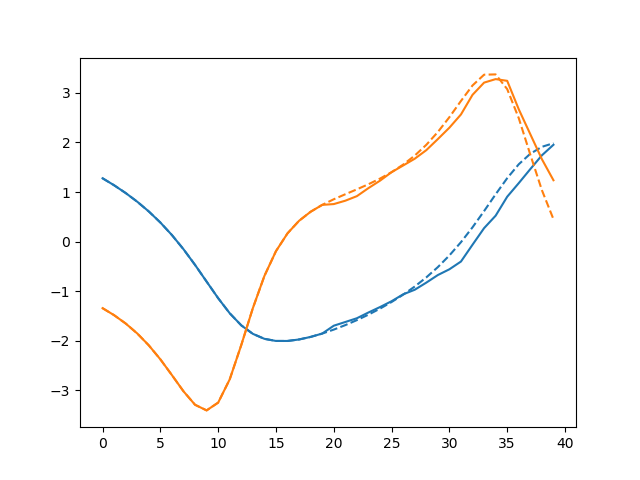}&
\includegraphics[width=9.5em]{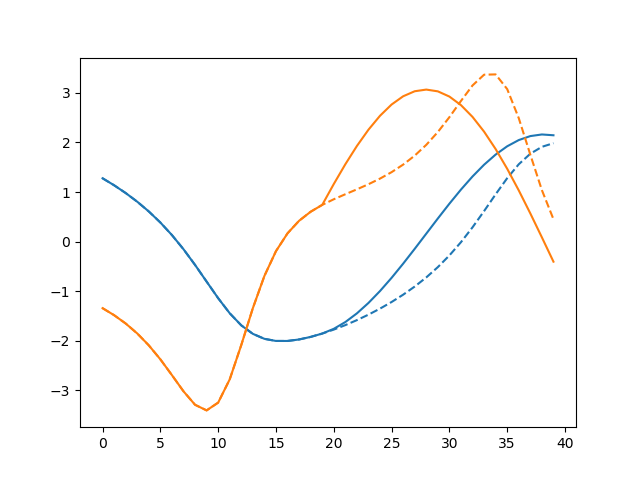}&
\includegraphics[width=9.5em]{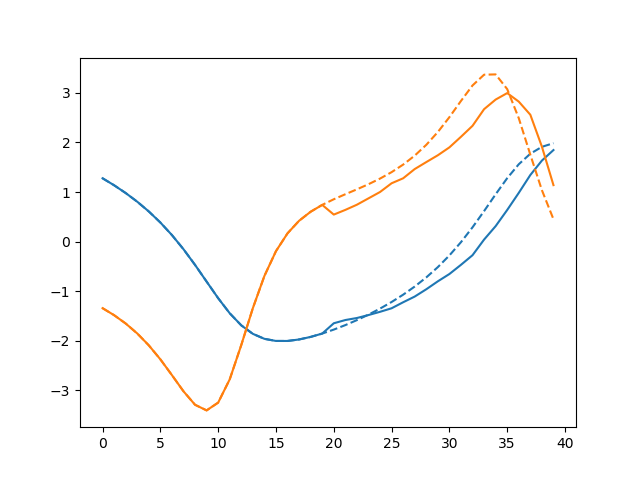}&
\includegraphics[width=9.5em]{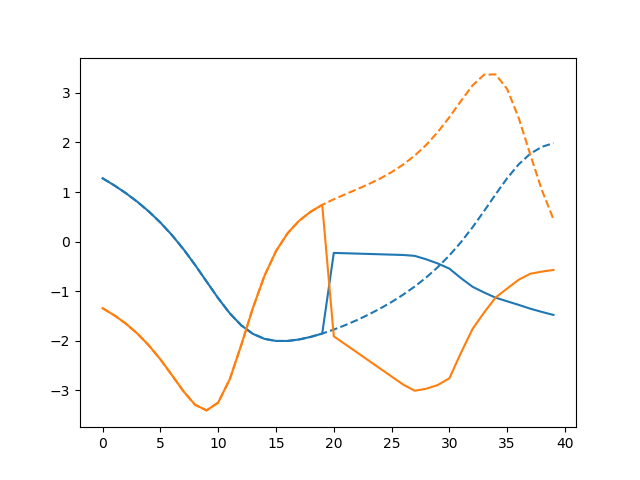}&
\includegraphics[width=9.5em]{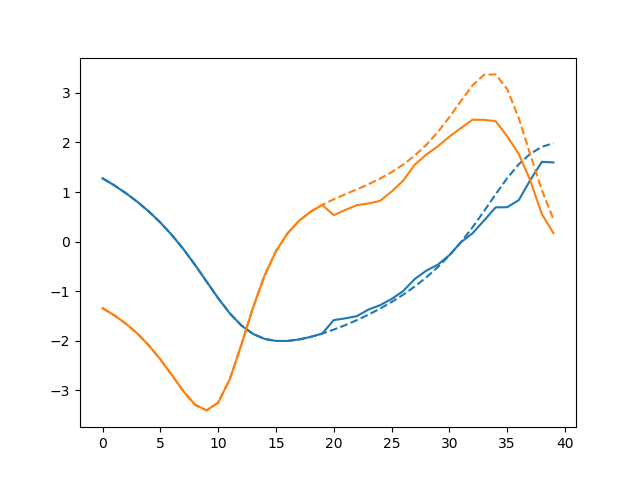}\\
(a) Ours & (b) DMD & (c) NDMD & (d) Finetune & (e) MAML \\
\end{tabular}}
\caption{Examples of future prediction with Van-der-Pol data. The dashed line is the true values, and the solid line is the predicted values. The first 20 timesteps were used as support time-series for predicting the next 20 timesteps. Each row shows the results of the same time-series by different methods.} 
\label{fig:predict}
\end{figure*}

\begin{figure*}[t!]
\centering
{\tabcolsep=0.3em \begin{tabular}{cccc}
\includegraphics[width=10em]{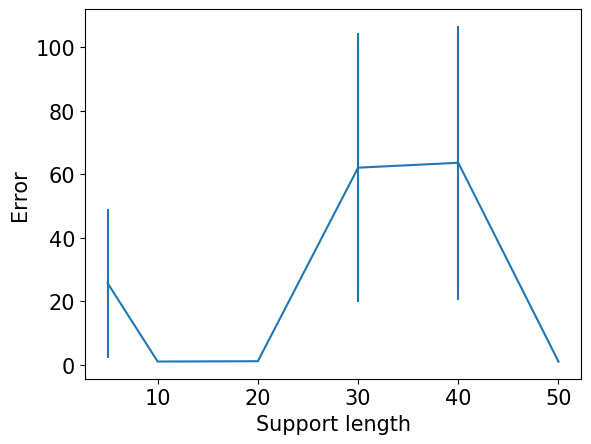}&
\includegraphics[width=10em]{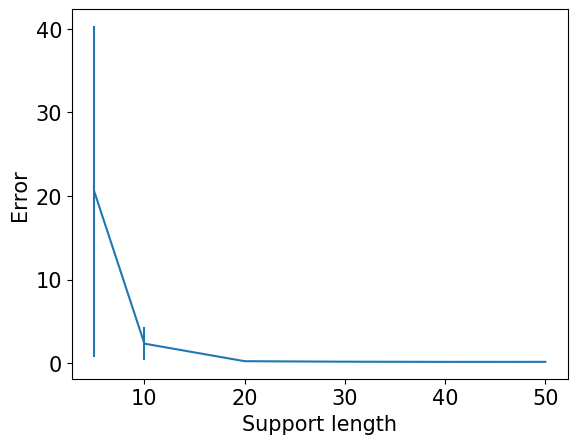}&
\includegraphics[width=10em]{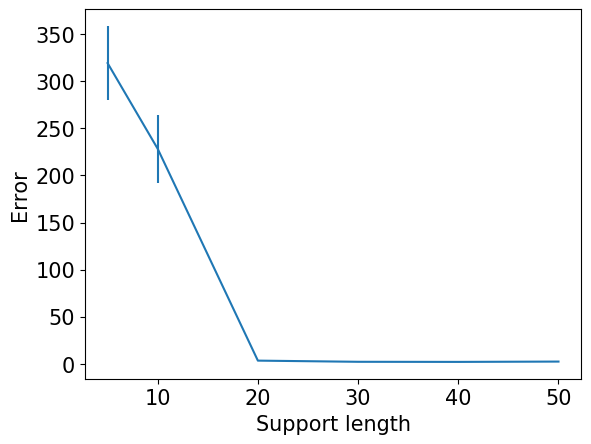}&
\includegraphics[width=10em]{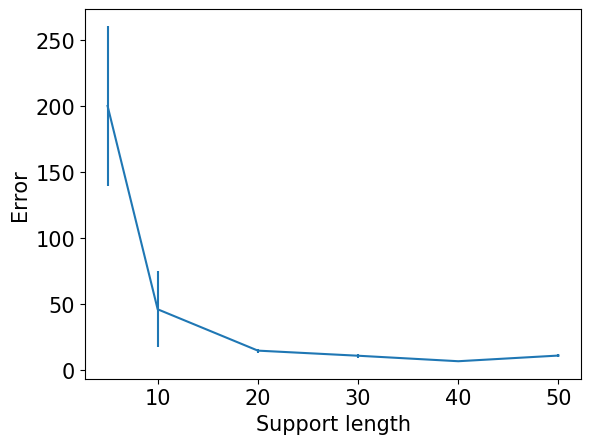}\\
(a) Synthetic & (b) Van-der-Pol & (c) Lorenz & (d) Cylinder-wake \\
\end{tabular}}
\caption{Prediction error with different lengths of support time-series.}
\label{fig:support_length}
\end{figure*}

\begin{figure*}[t!]
\centering
{\tabcolsep=0.3em \begin{tabular}{cccc}
\includegraphics[width=10em]{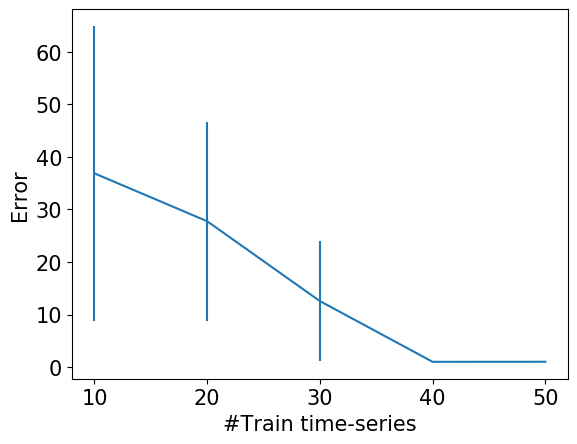}&
\includegraphics[width=10em]{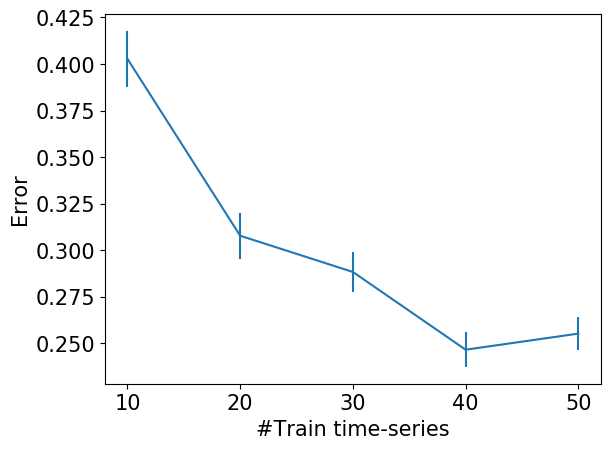}&
\includegraphics[width=10em]{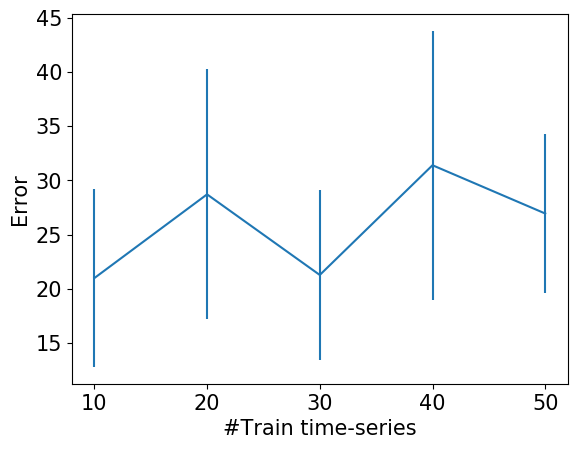}&
\includegraphics[width=10em]{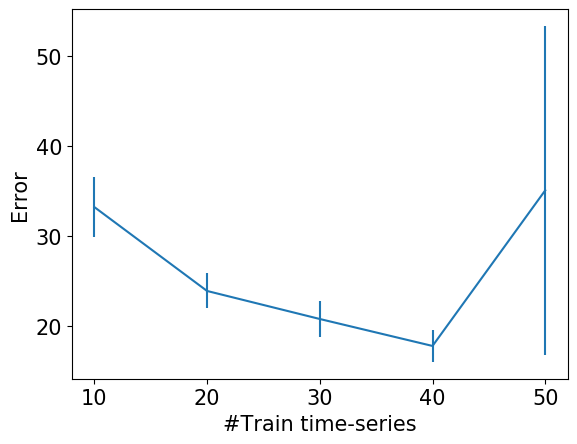}\\
(a) Synthetic & (b) Van-der-Pol & (c) Lorenz & (d) Cylinder-wake \\
\end{tabular}}
\caption{Prediction error with different numbers of training time-series.}
\label{fig:train_series}
\end{figure*}

\begin{table*}[t]
\centering
\begin{tabular}{lrrrrrrr}
\hline
& Ours & DMD & NDMD & Finetune & MAML & OursT & OursN\\
\hline
Synthetic & 7049 & - & 5552 & - & 64283 & 7596 & 5147\\
Van-der-Pol & 7785 & - & 6328 & - & 77228 & 8651 & 5686\\
Lorenz & 68934 & - & 54396 & - &  624310 & 72872 & 50531\\
Cylinder-wake & 8366 & - & 6663 & - & 71346 & 9331 & 5847\\
\hline
\end{tabular}
\\
(a) Training computational time
\\
\begin{tabular}{lrrrrrrr}
\hline
& Ours & DMD & NDMD & Finetune & MAML & OursT & OursN\\
\hline
Synthetic & 0.39 & 0.63 & 0.38 & 1.57 & 0.44 & 0.39 & 0.38\\
Van-der-Pol & 0.19 & 0.19 & 0.19 & 1.09 & 0.24 & 0.19 & 0.18\\
Lorenz & 0.19 & 0.21 & 0.19 & 1.12 & 0.20 & 0.18 & 0.19\\
Cylinder-wake & 0.28 & 0.36 & 0.28 & 1.34 & 0.29 & 0.28 & 0.28\\
\hline
\end{tabular}\\
(b) Test computational time 
\caption{(a) Training computational time in seconds. (b) Test computational time per a target time-series in seconds. There is no training phase with DMD since DMD does not have neural networks. Finetune uses the trained results of NDMD.}
\label{tab:time}
\end{table*}

We evaluated the eigenvalue estimation performance
using the Synthetic data
since the true eigenvalues can be calculated with the Synthetic data.
The true eigenvalues of the Koopman operator were
given by the eigenvalues of the hidden linear transition matrix.
The error between the estimated and
true eigenvalues was calculated by the mean of the minimum of the absolute error:
\begin{align}
    \frac{1}{2}\Bigg(\frac{1}{D}\sum_{d=1}^{D}\min_{d'}|\hat{\lambda}_{d}-\lambda_{d'}|+\frac{1}{D'}\sum_{d'=1}^{D'}\min_{d}|\hat{\lambda}_{d}-\lambda_{d'}|\Bigg),
\end{align}
where $\hat{\lambda}_{d}$ and $\lambda_{d}$ are estimated and true eigenvalues, and
$D$ and $D'$ are the numbers of estimated and true eigenvalues.
Table~\ref{tab:eigen} shows the result.
The proposed method achieved the lowest error.
This result indicates that the proposed method can estimate 
the eigenvalue of the Koopman operator
from short time-series by training using multiple related time-series.

We also evaluated the performance 
with the rooted mean squared error between the predicted and true future measurement
vectors.
Table~\ref{tab:rmse} shows the result.
The proposed method achieved the lowest prediction error with all data.
The prediction error by DMD was high 
since DMD estimated the Koopman matrix in the measurement space.
Although NDMD used multiple time-series for training,
since it learned common functions across all time-series 
and could not handle time-series-specific properties,
it failed to predict future measurement vectors.
On the other hand, the proposed method
adapted the functions for each time-series
using time-series representations inferred by the LSTM
while extracting the common useful knowledge by sharing the neural networks.
Finetune failed to improve the performance from NDMD with Van-der-Pol data.
It is because finetuning with the short time-series 
can deteriorate the performance by overfitting. 
The error by MAML was lower than that by Finetune since
MAML trained the neural networks such that they perform better when finetuned.
However, MAML performed worse than the proposed method.
MAML shared the initial parameters of the neural networks. 
If the properties of the dynamics are similar across time-series,
the shared initial parameters can be useful for transferring knowledge.
However, if the properties are drastically 
different across time-series,
it would be difficult to find good parameters for the diverse dynamics
from the shared initial parameters. 
In contrast, the proposed method can flexibly modify the mapping functions 
by inputting time-series representations into the neural networks,
which resulted in its better performance.
OursT was worse than the proposed method. This result indicates that 
minimizing the test prediction error is more effective than minimizing the training prediction error.
OursN was also worse than the proposed method, which implies 
the importance of time-series representations with the proposed method.

Figure~\ref{fig:predict} shows examples of future prediction with
Van-der-Pol data. 
Since DMD assumed linear dynamics in the measurement space,
its prediction gradually diverged from the true values.
With Finetune, the prediction drastically 
differed from the true values 
because it could not learn appropriate embedding functions
by overfitting to the short time-series.
Figure~\ref{fig:support_length} shows the prediction error with different lengths
of support time-series by the proposed method.
As the length increased, the error decreased.
It is because the time-series representations can be
estimated more properly with longer support time-series
by the bidirectional LSTM.
Figure~\ref{fig:train_series} shows the prediction error with different numbers
of time-series in training data.
As the number of training time-series increased,
the error decreased.
By training from many time-series,
we can increase knowledge that can be used 
for modeling unseen test time-series.
Table~\ref{tab:time}
shows the training and test computational time
by computers with 2.10GHz CPU.
The proposed method took a long training time, but 
it was shorter than MAML.
Since the proposed method do not need to retrain 
given a new short time-series,
its test time was short.

\section{Conclusion}
\label{sec:conclusion}

We proposed a neural network-based meta-learning method 
for Koopman spectral analysis.
Although we believe that our work is an important step
for analyzing nonlinear dynamical systems 
even when only short time-series are available,
we must extend our approach in several directions.
First, we plan to use the proposed method for model-based control of nonlinear dynamical systems.
Second, we will investigate different types of neural networks 
for extracting information from short time-series,
such as attentions~\cite{kim2019attentive,lee2019set}.

\bibliographystyle{abbrv}
\bibliography{main}
    
\end{document}